\documentclass[letterpaper, 10 pt, conference]{ieeeconf}
\IEEEoverridecommandlockouts
\usepackage{macros}
\usepackage{multicol}
\usepackage[bookmarks=true]{hyperref}

\newtheorem{deff}{Theorem}
\newtheorem{definition}[deff]{Definition}
\newtheorem{theo}{Theorem}
\newtheorem{theorem}[theo]{Theorem}

\begin{document}

\title{Real-time Funnel Generation
for Restricted Motion Planning}
\author{
Hadi Ravanbakhsh, Forrest Laine and Sanjit A. Seshia\\
\textit{Department of Electrical Engineering and Computer Science} \\
\textit{University of California, Berkeley, USA}
}

\maketitle

\begin{abstract}
In autonomous systems, a motion planner generates reference trajectories which are tracked by a low-level controller. For safe operation, the motion planner should account for inevitable controller tracking error  when generating avoidance trajectories. In this article we present a method for generating provably safe tracking error bounds, while reducing over-conservatism that exists in existing methods. 
We achieve this goal by restricting possible behaviors for the motion planner. We provide an algebraic method based on sum-of-squares programming to define restrictions on the motion planner and find small bounds on the tracking error.
We demonstrate our method on two case studies and show how we can integrate the method into already developed motion planning techniques. Results suggest that our method can provide acceptable tracking error wherein previous work were not applicable.
\end{abstract}
\section{Introduction}
Every dynamic robotic system that operates in non-static environments requires some ability to plan motions in real-time to interact with the world. Depending on the system in use, the application at hand, and the computational tools available, there are numerous different ways the motion planning problem can be formulated and solved. Typically constant among all of these formulations is their use of simplified models, and in practice they are equipped with a feedback control to account for deviation of the actual system from the planned trajectory.

Recent developments toward safe motion planning consider the low-level feedback control to provide formal guarantees on the system's ability to track the motion plans. In particular, the FaSTrak framework~\cite{FaSTrack,fastrak-sos} computes (a) a feedback law along and (b) an upper-bound on the deviation from the plan (tracking error). The system can use the feedback law to account for the deviation from the plan. The feedback law guarantees that the actual system remains inside a \emph{funnel} around the plan and does not deviate too much. This information can be used inside the motion planner to avoid unsafe behaviors~\cite{FaSTrack}.
However, this upper-bound on the tracking error can be large and impractical for certain types of systems as we show in this article. To address the issue and decrease the tracking error, we propose to restrict the motion planner.

We introduce a class of auxiliary functions, namely ``\names" which allow us to efficiently calculate feedback laws and funnels for a set of \emph{admissible} reference trajectories. Then, we require the motion planner to generate only trajectories that belong to the admissible set. We show that a smaller admissible set (i.e., a more restricted motion planner) yields a smaller tracking error.

We provide a method to systematically find \names, as well as set of admissible trajectories. Our method uses sum-of-squares (SOS) programming which in turn is reduced to semi-definite programming problems, making our solution efficient and relatively scalable.
 We demonstrate the effectiveness of our method on two case-studies: (a) an altered inverted pendulum and (b) a ground vehicle, and discuss how our method can be integrated into motion planning methods to generate certified motions.

In the next section, we provide background, introduce necessary notation, and formulate the problem. We then describe a motivating example wherein we show that without restricting the motion planner, the tracking error can get unacceptably large. In the fourth section, we introduce \names\ and show how we can use them inside motion planners. Finally, we consider two case-studies, compare with related work, and discuss future directions.

\section{Background}
We consider a dynamical systems of the form:
\[ \dot{\vx} = f(\vx, \vu) \,,\]
where $\vx \in \reals^n$ and $\vu \in U \subseteq \reals^m$ denote the state and control input of the system, and $f$ is a polynomial.
For an initial state $\vx_0$ and an input trajectory $\vu(\cdot)$, a state trajectory $\vx(\cdot)$ is uniquely defined under some mild assumptions~\cite{meiss2007differential}.
Given $\vx_0$, motion planning is used to generate an open loop reference control signal corresponding to some reference desired trajectory $\vx_r(\cdot)$. In practice, however, the trajectory of the system as it executes the reference control may not match the desired trajectory due to disturbances or unmodeled effects. Typically, a feedback law is used to modify the reference control in order to keep states near the reference trajectory. 
Assume $\vx_r(\cdot)$ and $\vu_r(\cdot)$ are given over a finite time horizon $T$. Then a feedback law $\ctrl : [0, T] \times X \rightarrow U$ maps a finite-time interval $\Tau: [0, T]$ and state $\vx$ to feedback $\vu$. Simple methods include PID or LQR for a linearization of the system dynamics around the reference trajectory.
While this assumption separates reference generation from feedback generation~\cite{frazzoli2002real}, nonlinearities may cause unexpected results and ideally, one wishes to know the region around the reference in which the feedback law is effective.

To formally study the effect of the feedback law $\ctrl$, we consider regions of finite-time invariant (funnels) around $\vx_r(\cdot)$. A funnel guarantees a state that is inside the funnel (close to $\vx_r(\cdot)$), remains inside the funnel for a finite time. 
Funnels could be used to provide guarantees when the actual system is subject to disturbances~\cite{majumdar2017}.
Funnels are also powerful tools to reason about \emph{safety} and \emph{reachability}. 
If a funnel is inside a safe region, as long as the state is inside the funnel, the system remains safe until time $T$. 
Moreover, if a system starts inside the funnel, the system will reach the tail of the funnel, a property suitable for state space exploration~\cite{tedrake2010lqr}.

\begin{figure}[t]
\vspace{0.2cm}
\begin{center}
	\includegraphics[width=0.25\textwidth]{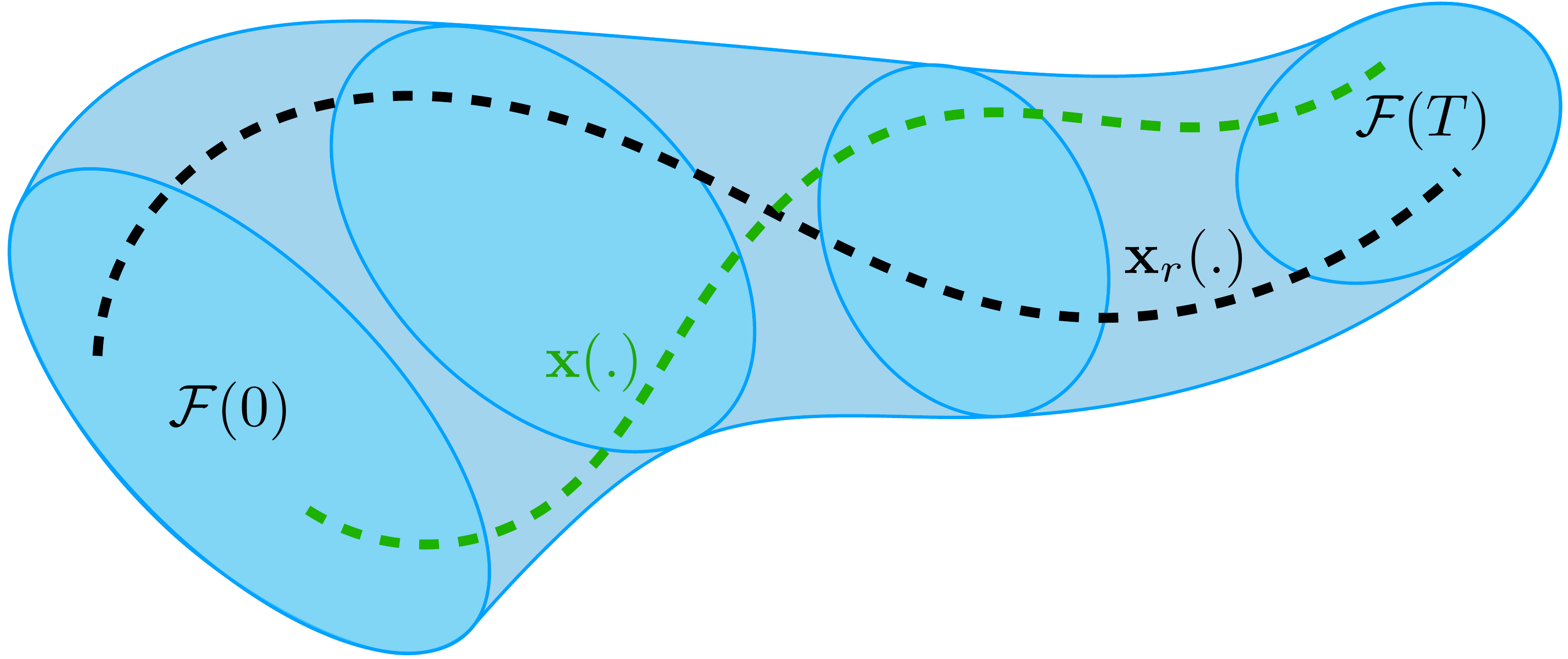}
\end{center}
\caption{A Schematic View of a funnel covering a reference.}\label{fig:funnel} 
\end{figure}

In this article, we are interested in generating reference trajectories along with (i) feedback laws and (ii) corresponding funnels around the reference at runtime.
The reference state and reference input are represented as $\vx_r$ ($\in X_r$) and $\vu_r$ ($\in U_r$), respectively. 
In this article, we assume $X = X_r$ and $U = U_r$ and we leave the extension to cases where $X_r \neq X$ for future work (e.g., see~\cite{FaSTrack}).
For a given closed-loop system, a funnel $\F : \Tau \rightarrow P(X)$ is a time-varying set s.t. $(\forall t \in \Tau) \ \vx(t) \in \F(t) \implies (\forall t' \in [t,T]), \ \vx(t') \in \F(t')$,
meaning that if a state is inside the funnel, it will remain in the funnel until time $T$. A funnel covers a reference $\vx_r(\cdot)$ if $\vx_r(t) \in \F(t)$ for all times (Fig.~\ref{fig:funnel}).

Now, consider a motion planning problem where generating the reference at runtime is the goal. Ideally, we wish to have a higher order function, namely a \emph{funnel generator} $\G$ that maps a reference ($\Tau \rightarrow X_r$) to a corresponding feedback law ($\Tau \times X \rightarrow U$) and a funnel ($\Tau \rightarrow P(X)$) covering the reference.

Majumdar et al.~\cite{majumdar2017} restrict the domain of a funnel generator $\G$ to be finite and provide a library of nominal trajectories, each with a separate feedback law designed for them. Then, at runtime, the motion planner selects a proper nominal trajectory from the library. While this method allows for small tracking errors (optimal tracking), the local motion planner is not flexible and must choose a trajectory from a finite library. \looseness = -1

Herbert et al.~\cite{FaSTrack} assume a reference trajectory is controlled by an adversary (a motion planner), and a funnel  generator is designed according to the worst-case behavior of the reference. Then, the motion planner can generate any arbitrary trajectory and the funnel corresponding to that trajectory is efficiently obtained at runtime. In this case, the planner has maximum flexibility, but generated funnels can be conservative (large tracking error) as they consider all the possible behaviors the motion planner can produce. 
To mitigate this issue, in this article we extend this method by restricting the motion planner to only generate reference trajectories in a subset of the trajectory space, called admissible set. The tracking error analysis is then done over the admissible set, allowing for a more refined tracking bound. 
For example, tracking error bounds for a car driving with high speed might be much smaller than the tracking error bound computed for references which are allowed to make arbitrary stops.

\section{Restricting the Planner}
To narrow the gap between an inflexible motion planner with optimal tracking error~\cite{majumdar2017} and a fully flexible motion planner with a large tracking error~\cite{FaSTrack}, we provide a solution that yields a trade-off between flexibility and optimality. 
Our goal is to find a feedback law that can generate proper behaviors for a class of reference trajectories. Then, similar to previous work~\cite{majumdar2012,fridovich2018planning}, we can have multiple funnel generators to cover all possible motions. In the next step, we need to design a motion planner that can use these funnel generators to produce and concatenate funnels to implement a robust plan.

Let $\X$ be the set of all admissible reference trajectories. We are interested in a funnel generator with domain $\X$. 
Here, we provide a representation for $\X$, suitable for realtime reference generation. 
The dynamics of the reference is described by $\dot{\vx_r} = f_r(\vx_r, \vu_r)$,
noting the reference can have a different dynamics than the system. This accounts for perhaps a simpler model which is used for efficient motion planning. For brevity, let $\vr = [\vx_r, \vu_r]$, which belongs to some set $R : X_r \times U_r$.
Formally, we define
$\X : \{ \vx(\cdot) \ | \ \vr(t) \in R^s(t) \subseteq R \}$, 
where $R^s$  is a time-varying set. 

\begin{figure}[t]
\vspace{0.2cm}
\begin{center}
	\includegraphics[width=0.35\textwidth]{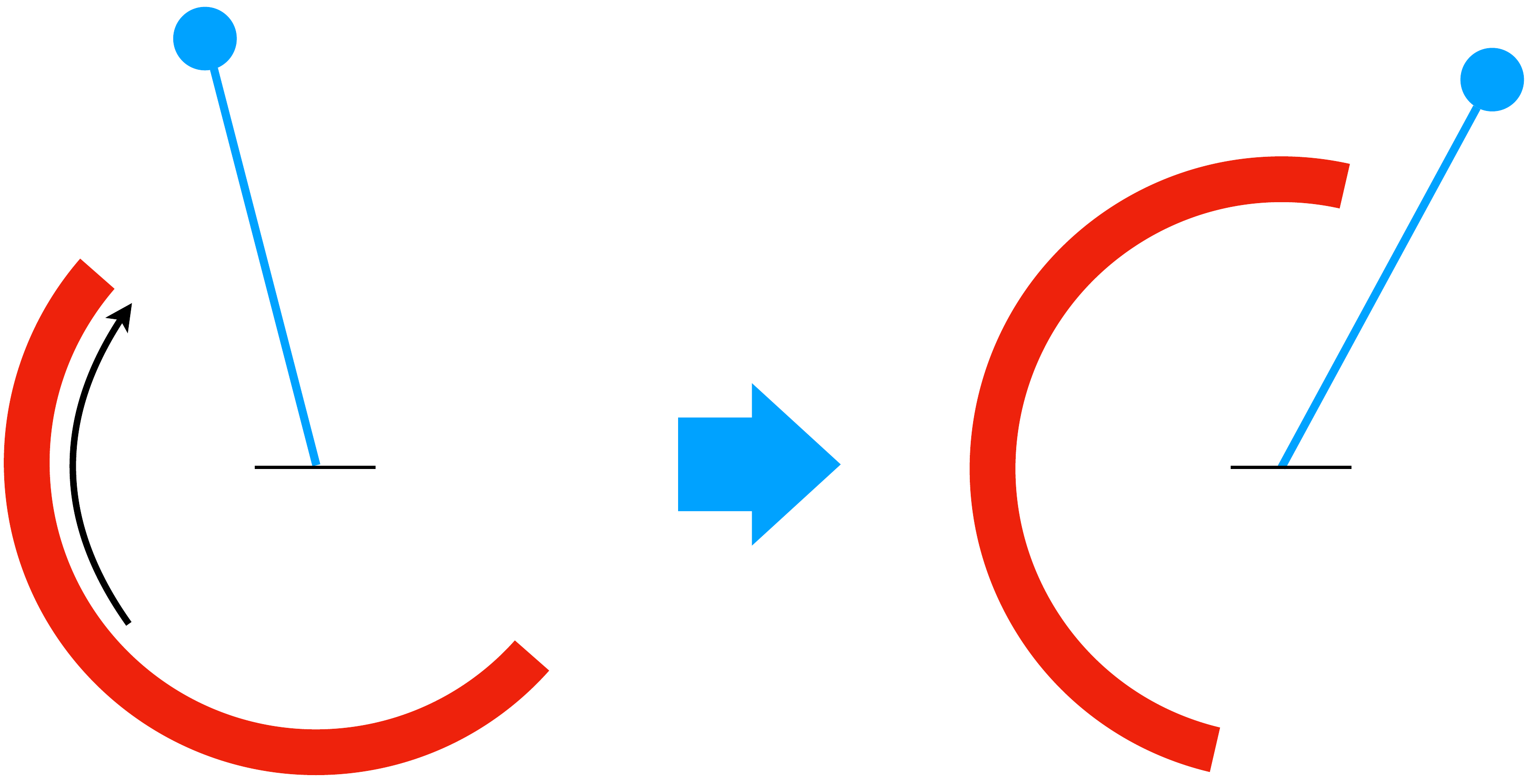}
\end{center}
\caption{A Schematic View of Inverted Pendulum with Moving Obstacles.}\label{fig:inverted-pendulum-view} 
\end{figure}

\paragraph{An Illustrative Example}
Consider a system similar to an inverted pendulum with dynamics given by
\[
\dot{\theta} = \omega {\color{black}- 0.2sin(\theta)} \ , \ \dot{\omega} = sin(\theta) + u \,,
\]
where 
$|u| \leq 1.15$ is a control input. 
A schematic view of such system is shown in Fig.~\ref{fig:inverted-pendulum-view}. As depicted, there are moving obstacles and we need a motion planner to avoid them. 
We compute a linearized version of this model to be used by the motion planner, wherein dynamics are given by $\dot{\theta_r} = \omega_r$, $\dot{\omega_r} = u_r$, and $u_r$ is the control for the reference/planning model. The goal of the tracker is to always keep true state of the system near the reference:
\[
	|\theta(t) - \theta_r(t)| < 0.2 \, , \, |\omega(t) - \omega_r(t)| < 0.2\,.
\]

It is easy to show that if $|u_r| < 0.7$, such tracking is not possible for all reference trajectories since for large values of $\theta$, the difference between the actual system and the reference gets large. However, as we will show later in this article, if we restrict the reference s.t. for all $t$, $|\theta_r(t)| < 0.5$, we can in fact design a feedback law to address the tracking problem. This fact motivates us to extend FaSTrak~\cite{FaSTrack,fastrak-sos} to design task-specific funnel generators. In other words, the motion planner can generate only track-able trajectories by considering some constraints on the reference.

\section{Designing Funnel Generators}
In this section, we provide an algebraic method to design funnel generators. Before discussing our approach, we provide some background on how to define funnels in algebraic forms.

\begin{definition}[Funnel Function~\cite{TOBENKIN2011}]
Given a reference $\vx_r(\cdot)$, and a feedback law $\ctrl$, a funnel function $V$ is a smooth radially unbounded function $V: \Tau \times X \rightarrow \reals^+$ s.t. $\forall t \in \Tau$
\begin{align}
	(\vx = \vx_r(t)) = 0 \implies & V(t, \vx) < \beta \\
	V(t, \vx) = \beta \implies & \frac{dV}{dt}  < 0 \,,
\end{align}	
where $\frac{dV}{dt} = \nabla_{\vx} V(t, \vx) \cdot f(\vx, \ctrl(t, \vx)) + \frac{\partial V}{\partial t}(t, \vx)$.
\end{definition}
A funnel function $V$ can be used to describe a funnel $\F : \Tau \rightarrow P(X)$ where 
\begin{equation}\label{eq:funnel}
\F(t) : \{\vx \ | \ V(t, \vx) \leq \beta\}\,.
\end{equation}
The first condition makes sure that the region in which the actual system and the reference are identical ($\vx = \vx_r(t)$) lies in the interior of funnel $\F$. The second condition guarantees that when the state reaches the boundary of $\F$, it is pushed back inside that region until time $T$. These conditions guarantee that $\F$ defined in Eq.~\eqref{eq:funnel} is a funnel.
Going one step further, one can design a feedback law $\ctrl$ and a funnel function $V$ at the same time using algebraic methods~\cite{TOBENKIN2011}.  
 Moreover, this technique can handle input saturation and disturbances as well~\cite{majumdar2017}. However, for brevity, in the rest of this section we consider systems without disturbances and their extension to systems with disturbances is straightforward (cf.~\cite{majumdar2017}).

\subsection{\Names}
One of the main limitations of funnel functions is that they assume a single fixed reference trajectory.
To allow variable references, we introduce auxiliary function $\hat{V}$ (\name), suitable for defining funnel generators. This auxiliary function is mapped to a funnel function $V$ after a reference trajectory in $R^s(\cdot)$ is generated. The computation of $\hat{V}$ is carried out offline and the final funnel (as well as the feedback law) is generated at runtime. 

 \begin{definition}[\NAME]
	A \name\ w.r.t. a region $R^s(\cdot)$ is a smooth radially unbounded function $\hat{V} : \Tau \times X \times X_r \rightarrow \reals$ s.t. for all $t \in \Tau$
	\begin{equation}\label{eq:meta-cert}
	\begin{array}{rl}
	(\vx = \vx_r) \implies \hspace{-0.4cm} & \hat{V}(t, \vx, \vx_r) < \beta \\
	\hspace{-0.6cm}	(\forall \vr \in R^s(t)) & \hat{V}(t, \vx, \vx_r)=\beta\implies \hspace{-0.1cm} \frac{dV}{dt} < 0 ,
	\end{array}
	\end{equation}
where $\frac{d\hat{V}}{dt} = \nabla_{\vx} \hat{V}(t, \vx, \vx_r) \cdot f(\vx, \hat{\ctrl}(t, \vx, \vx_r, \vu_r)) +  \nabla_{\vx_r} \hat{V}(t, \vx, \vx_r) \cdot f_r(\vx_r, \vu_r) +  \frac{\partial \hat{V}}{\partial t}(t, \vx, \vx_r) < 0$, and $\hat{\ctrl}: \Tau \times X \times X_r \times U_r \rightarrow U$.
\end{definition}

\begin{figure}[t]
\vspace{0.2cm}
\begin{center}
	\includegraphics[width=0.45\textwidth]{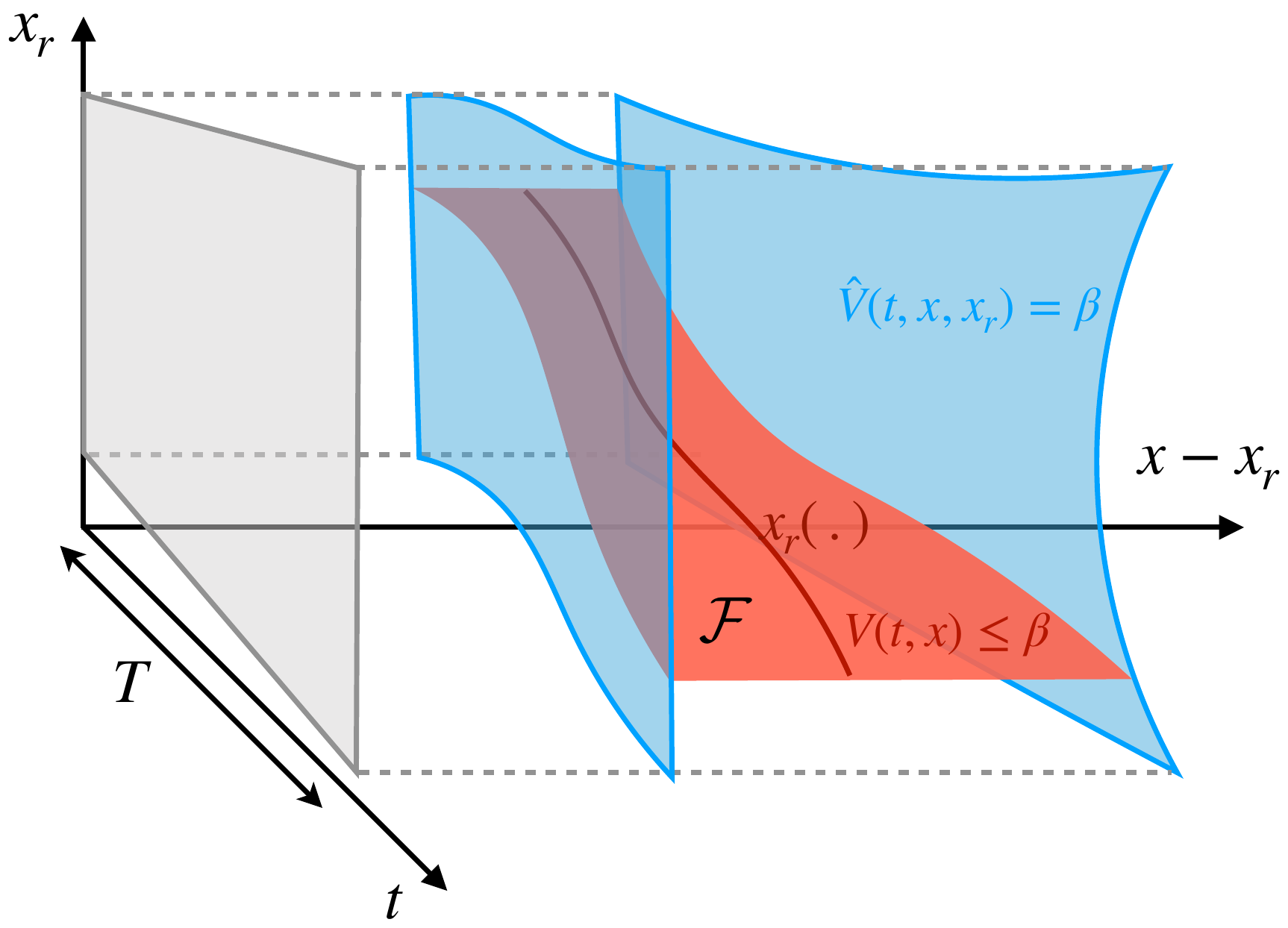}
\end{center}
\caption{A Schematic View of \Name. The red surface is the funnel and the blue surface shows the boundary of $\hat{V}(t, \vx, \vx_r) \leq \beta$.}\label{fig:fgf} 
\end{figure}

Notice that in addition to $\vx$ and $t$, $\hat{V}$ ($\hat{\ctrl}$) is defined over $\vx_r$ ($\vx_r$ and $\vu_r$), as the reference is not fixed. As shown in Fig.~\ref{fig:fgf}, having a function $\hat{V}$, first a reference in $R_r^s(\cdot)$ is generated. Then, using $\hat{V}$, one can have a funnel around the reference (red surface) in closed form.
Formally, for a given reference $\vr(\cdot): [\vx_r(\cdot), \vu_r(\cdot)]$, we get funnel $\F$ and feedback law $\ctrl$ as follows:
\begin{align}
\ctrl(t, \vx) &: \hat{\ctrl}(t, \vx, \vx_r(t), \vu_r(t))\, \label{eq:feedback-law-2} \\
\F(t) &: \{\vx \ | \ \hat{V}(t, \vx, \vx_r(t)) \leq \beta\} \,.\label{eq:funnel-2}
\end{align}

Using the feedback law in Eq.~\eqref{eq:feedback-law-2} and from definition of \name\ we conclude $\forall t \in \Tau$, $\hat{V}(t, \vx, \vx_r(t)) = \beta \implies \ \frac{d\hat{V}}{dt} < 0$,
which means $\F$ defined in Eq.~\eqref{eq:funnel-2} is a funnel that covers $\vx_r(\cdot)$.
\begin{theorem}
	Given a \name\ $\hat{V}$, as long as $\vr(\cdot) \in R^s$ and the feedback law is defined according to Eq.~\eqref{eq:feedback-law-2}, set $\F$ defined in Eq.~\eqref{eq:funnel-2} is a region of finite-time invariant around the reference trajectory.
\end{theorem}

\subsection{Motion Planning with Funnel Generators}
While planning robust constraint-satisfying trajectories is difficult in any setting, we believe the method presented here can help ease the difficulty of this task.
In this work, we considered motion planner as a black box and any kind of motion planning technique could be used so long as the generated reference respects the constraints. Moreover, if the motion planner discretizes the time, the funnels and thus all guarantees would be hold only for discrete times (cf.~\cite{majumdar2017}).

For a given funnel generator $\G$, a motion planner first generates the reference trajectory in the domain of $\G$. Then, the funnel generator yields the feedback-law for the trajectory in a closed form. However, considering an obstacle avoidance scenario, the motion planner should make sure not only the reference but also the funnel around it avoids obstacles. Depending on the motion planner, two methods could be used. A simple approach is to generate the reference and reject the solution if the funnel intersects unsafe region. Another solution is to encode the funnel (as a semi-algebraic set) inside the motion planning problem, and the funnels are used to `pad' the obstacles by the width of the funnel, which can be a function of the reference.

\subsection{Searching for \NAMES}
Previous work~\cite{TOBENKIN2011} shows how we can find a funnel function and the feedback law using sum-of-squares programming~\cite{Parrilo2003}.
To extend this method to find a \name, in addition to $\hat{V}$ and $\hat{\ctrl}$, we need to find region $R^s(\cdot)$ for which $\hat{V}$ respects the constraints. 
From motion planning perspective, we prefer to define $R^s(\cdot)$ using a set of semi-algebraic (and convex) sets: $r_1(t, \vr) \geq 0, \ldots, r_{N_R}(t, \vr) \geq 0$.
However, from a verification perspective, these sets are polynomials of up to a fixed degree and we can only under-approximate the maximum region. 
Therefore, there are many maximal solutions and selecting one would be a design choice. Considering the efficiency of trajectory optimization, here we simply search for regions where in $R^s(\cdot)$ is a time-varying box. The shape of such a box we wish to maximize can be a design choice.

We provide a simple, yet effective solution. 
Starting with $ R^s(\cdot): \{ \vr^*(\cdot) \}$ (for a nominal trajectory $\vr^*(\cdot)$), we gradually expand the set.
For a set $S$ and a scalar $p \geq 0$, let $pS : \{p \vs \ | \ \vs \in S\}$. 
We parameterize $R^s(\cdot)$ using a scalar $p$:
\[R^{s}_p : (\{\vr^*(\cdot)\} \oplus p R^{expand}(\cdot)) \cap R^{max}(\cdot)\,,\]
where $\oplus$ is the Minkowski sum, $R^{expand}(t) \ni \vzero$ and $R^{max}(t)$ are design choices. Intuitively, by maximizing $p$, for all times, $R^{s}_p(t)$ is expanded along each direction in $R^{expand}(t)$ until it reaches boundary of $R^{max}(t)$ in that direction. To maximize $R^{s}_p(t)$, we simply maximize $p$.

Assuming all sets are basic semi-algebraic sets, the problem of maximizing $p$ can be solved by simply performing a line search over $p$ (starting from $p = 0$). 
More precisely, for a fixed $p$, region $R^s(\cdot)$ is defined and our goal is to find $\hat{V}$ and $\hat{\ctrl}$ s.t. $\frac{dV}{dt} < 0$. Similar to previous work~\cite{TOBENKIN2011}, starting from an initial guess $\hat{V}$, we try to minimize $\gamma$ s.t. $\frac{d\hat{V}}{dt} < \gamma$ by repeating the following steps:
\begin{itemize}
	\item minimize $\gamma$ by fixing $\hat{V}$ and searching for a $\hat{\ctrl}$
	\item minimize $\gamma$ by fixing $\hat{\ctrl}$ and searching for a $\hat{V}$
\end{itemize}
Assuming all sets are basic semi-algebraic sets and functions are polynomial, each step is solved using a sum-of-squares optimization. Once $\gamma < 0$, $\hat{V}$ and $\hat{\ctrl}$ are solutions for $R^s_p(\cdot)$.

\section{Experiments}
In this section we review some motion planning algorithms through two case-studies. For each case-study first we design funnel generators, and then we discuss how they can be integrated into motion planning techniques. 

\subsection{Case Study I: Inverted Pendulum}
Going back to the illustrating example, in order to design funnel generators we start with a funnel function $V$ for a fixed reference $\theta_r(t) = \omega_r(t) = 0$.
By choosing proper $R^{max}(.)$ and $R^{expand}(.)$ :
\begin{align*}
\begin{array}{ll}
 R^{max}:
 \theta_r \in [-0.5, 0.5], \omega_r \in [-1, 1] , u_r \in [-1, 1] \\
R^{expand}:
  \theta_r \in [-10, 10], \omega_r \in [-10, 10] , u_r \in [-1, 1]\,,
\end{array}
\end{align*}
we find the following maximal region $R^s(\cdot)$:
\[
\theta_r(t) \in [-0.5, 0.5] \, , \, \omega_r(t) \in [-1, 1] \, , \, u_r(t) \in [-0.7, 0.7]\,.
\]
In other words, the system can track the reference when $\theta_r$ is close to origin and moves slowly. This yields our first funnel generator. 
Also, if we use 
\[
\begin{array}{ll}
 R^{max}:
 \theta_r \in [-0.9, 0.9], \omega_r \in [-1, 1] , u_r \in [-1, 1]
 \end{array}
\]
the maximal region would be different:
\[
\theta_r(t) \in [-0.9, 0.9] \, , \, \omega_r(t) \in [-1, 1] \, , \, u_r(t) \in [-0.35, 0.35]\,.
\]
Intuitively, the reference can safely get further from origin (larger range for $\theta_r$) as long as it does not move rapidly (smaller range for $u_r$). These two regions yield two funnel generators to be used under different settings. Next, we consider the motion planning problem.

\begin{figure}[t!]
\vspace{0.2cm}
\begin{center}
	\includegraphics[width=0.4\textwidth]{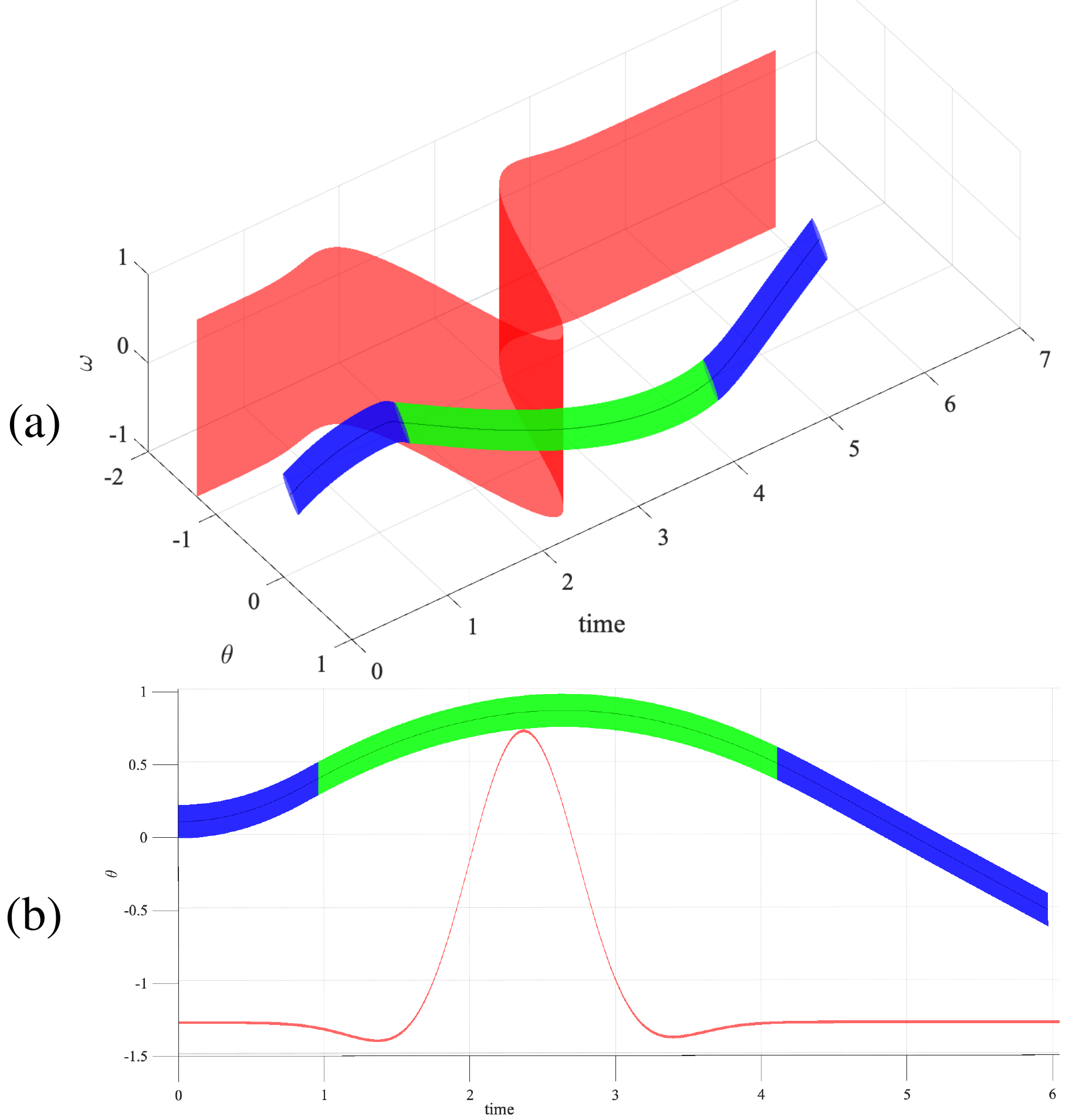}
\end{center}
\caption{Two Schematic Views of Concatenated Funnels for Inverted Pendulum}\label{fig:inverted-pendulum-funnels} 
\end{figure}

We consider a time-varying obstacle for $\theta$. The obstacle covers $\theta \leq 0.75$ at some point. None of funnel generators alone could yield a solution for the problem and we need to plan using multiple funnel generators. While planning using a single funnel generator is quite efficient in our experiments, optimally planning using multiple funnel generators requires a combinatorial search over sequences of funnels and we should make sure we can concatenate those funnels in switching times (cf.~\cite{majumdar2017}). While this induces an extra layer of complexity in the problem, there are in practice only few funnel generators which need to be considered to get acceptable tracking bounds, making the search space much smaller. 
For this case-study, we assume the sequence of funnel generators is already known.
Knowing funnel orders, generating the reference can be formulated as a single trajectory optimization. Figure \ref{fig:inverted-pendulum-funnels} illustrates the result of such a planning problem in the pendulum domain. It is seen how these different funnels can be strung together to plan an evasive maneuver which avoids the obstacle (represented red, as some lower bound on the value of $\theta$), which requires planning a trajectory that traverses into $0.5 \leq |\theta_r|$  region (using the green funnel) before returning.

\subsection{Ground Vehicle}
Consider a bicycle model of a ground vehicle where the state of the vehicle is modeled with its position $x$, $y$, orientation $\theta$, and speed $v$ ($\vx :[x, y, \theta, v]^t$). The control inputs are thrust $u_1 \in [-5, 5]$ and steering $u_2 \in [-1, 1]$ ($\vu:[u_1, u_2]^t$). To model the lateral slide, we use a disturbance $d$ in the following dynamics:
\begin{align*}
\dot{x} &= v \sin(\theta) + v \cos(\theta) d \ &, \ \dot{y} &= v \cos(\theta) + v \sin(\theta) d \\
	\dot{\theta} &= \frac{v}{l} u_2 \ &, \ \dot{v} &= u_1\,,
\end{align*}
where $l = 0.3$, $u_1 \in [-1, 1]$, $u_2 \in [-5, 5]$ and $d \in [-0.05, 0.05]$.
For the reference generation, we set $d = 0$ and use a simplified model. Without restricting the motion planner, we could only generate funnels that rapidly expand in a second (not useful for our motion planning problems). To restrict the motion planner, we consider three different tasks for which we wish to design a funnel generator: (i) \emph{Maneuvering} when the speed is around $v = 1.5$, (ii) \emph{Stopping} the vehicle, (iii) \emph{Accelerating} a stopped vehicle.

\begin{figure}[t]
 \vspace{0.2cm}
\begin{center}
	\includegraphics[width=0.38\textwidth]{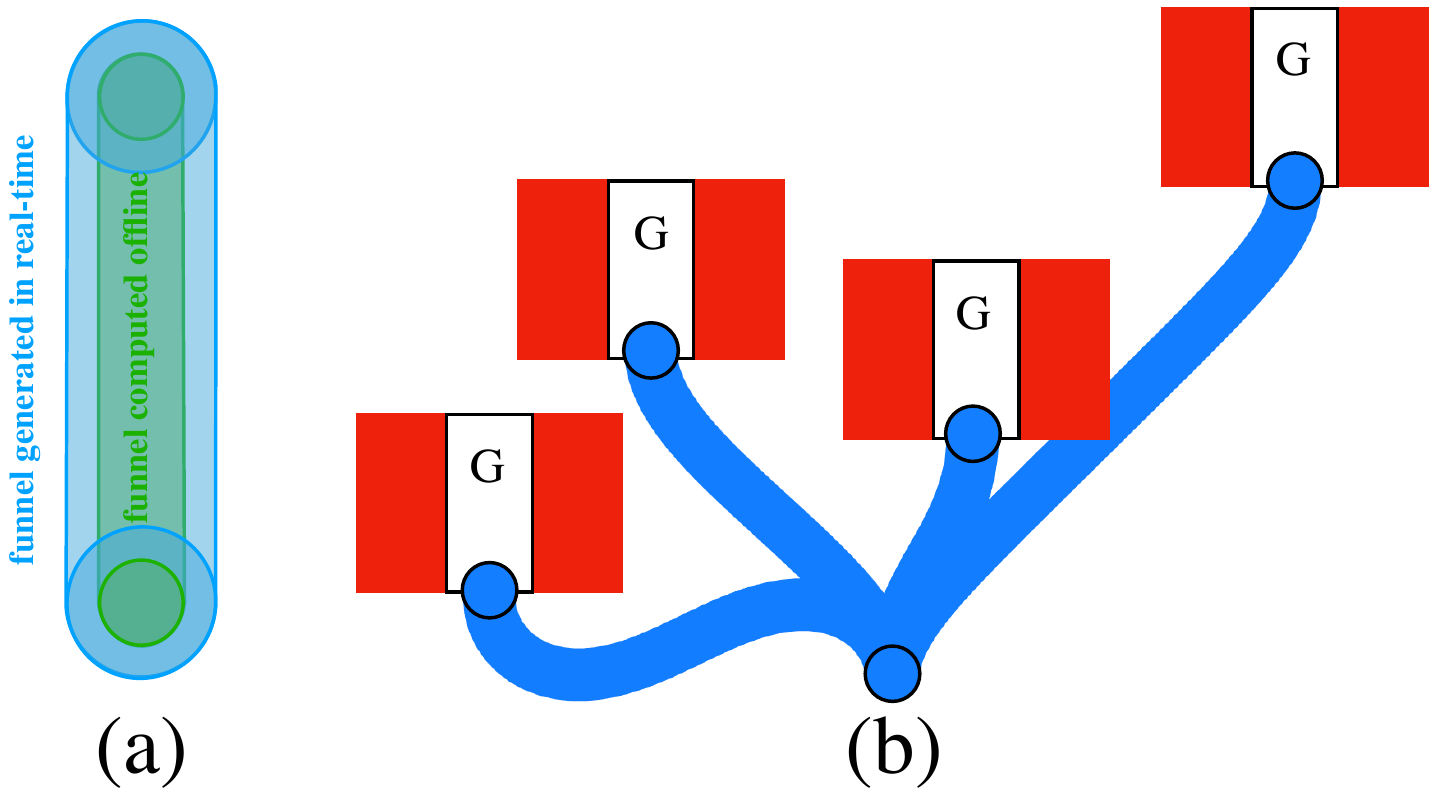}
\end{center}
\caption{Comparison with off-line funnel generation. (a) compares projection of funnels on $x$-$y$ axes. Blue funnel is  generated at runtime and the green one is generated off-line for a fixed trajectory (moving in a straight line). (b) shows different funnels generated at runtime for different goals.}\label{fig:lane-change} 
\end{figure}

\paragraph{Task I}
For the maneuvering task first we calculate a funnel function $V$ for a nominal behavior. 
By choosing proper $R^{max}$ and $R^{expand}$ we search for the largest range for $u_{r2}$ when range of $v_r$ and $u_{r1}$ are restricted by $R^{max}$. The results suggest that $V$ is a funnel generator function when $u_{r2}(t) \in [-0.6, 0.6]$, $u_{r1}(t) \in [-1, 1]$ and $v(t) \in [0.5, 2.5]$. A schematic view of funnels that are generate with this funnel generator is shown in Fig.~\ref{fig:lane-change}(b). As depicted, this funnel generator can generate a variety of funnels, suitable to perform a specific task, such as lane-change.

\paragraph{A Comparison with Funnel Library}
As mentioned earlier, when we consider a finite funnel library, we can minimize the tracking error as references are fixed. Fig.~\ref{fig:lane-change}(a) shows the funnel generated for a fixed reference trajectory (vehicle moving on a straight line), which is smaller than the one generated by a funnel generator.
On the down side, motion planner looses flexibility to deal with different environments. As demonstrated by Majumdar et al.~\cite{majumdar2017}, small libraries are sufficient for fast reactions. However, as the planning horizon increases, we need to add more funnels to the library.
For a comparison, we consider problem of moving from current position $x = 0$, $y = 0$ with velocity $v \in [0.5, 2.5]$ to a target region $G$ with velocity $v^* = 1.5$. The target region of size $1 \times 2$ is placed between two obstacles, but its position as well as initial speed of the vehicle are unknown until run-time. Assuming center of $G$ can be in range $[-5, 5] \times [1, 6]$, the task is to reach $G$ from below without collision. Our single funnel generator along with a motion planner can safely generate a reference for all possible cases (Fig.~\ref{fig:lane-change}(b)). However, we need at least $200$ funnels to safely solve the problem for all cases using a funnel library. We also note that if there is a narrow passage, funnels generated by a funnel generator maybe too large to provide safety guarantees.

\paragraph{Other Tasks}
For the stopping task, first we consider a nominal trajectory where the vehicle stops starting from $v = 1.5$. More precisely $v_r(t) : 1.5 (1-t)$ where $T = 1$. 

By choosing proper $R^{max}$ and $R^{expand}$ 
we find that $V$ is a funnel generator function if $v(t) \in [1.0, 2.0](1-t)$, $u_{r1}(t) \in [-2.0 , -1.0]$, and  $u_{r2}(t) \in [-0.5, 0.5](1-t)$.
These results suggest that steering input is more restricted when the AV is moving in a lower speed and the motion planner should consider these constraints to avoid generating non-trackable references.
Similar method can be used to compute a funnel generator function for accelerating tasks.

\paragraph{Motion Planning with Waypoints} 
One approach for motion planning is to generate a set of waypoints and then use a trajectory optimization technique (as a local motion planner) to connect them. In particular, we can move backward from the goal and at each step find a funnel containing a waypoint that fit into the next funnel (containing the next waypoint) until reaching the initial state. Then, using the corresponding feedback laws, the plan is robustly executed.

\begin{figure}[t]
\vspace{0.2cm}
\begin{center}
	\includegraphics[width=0.3\textwidth]{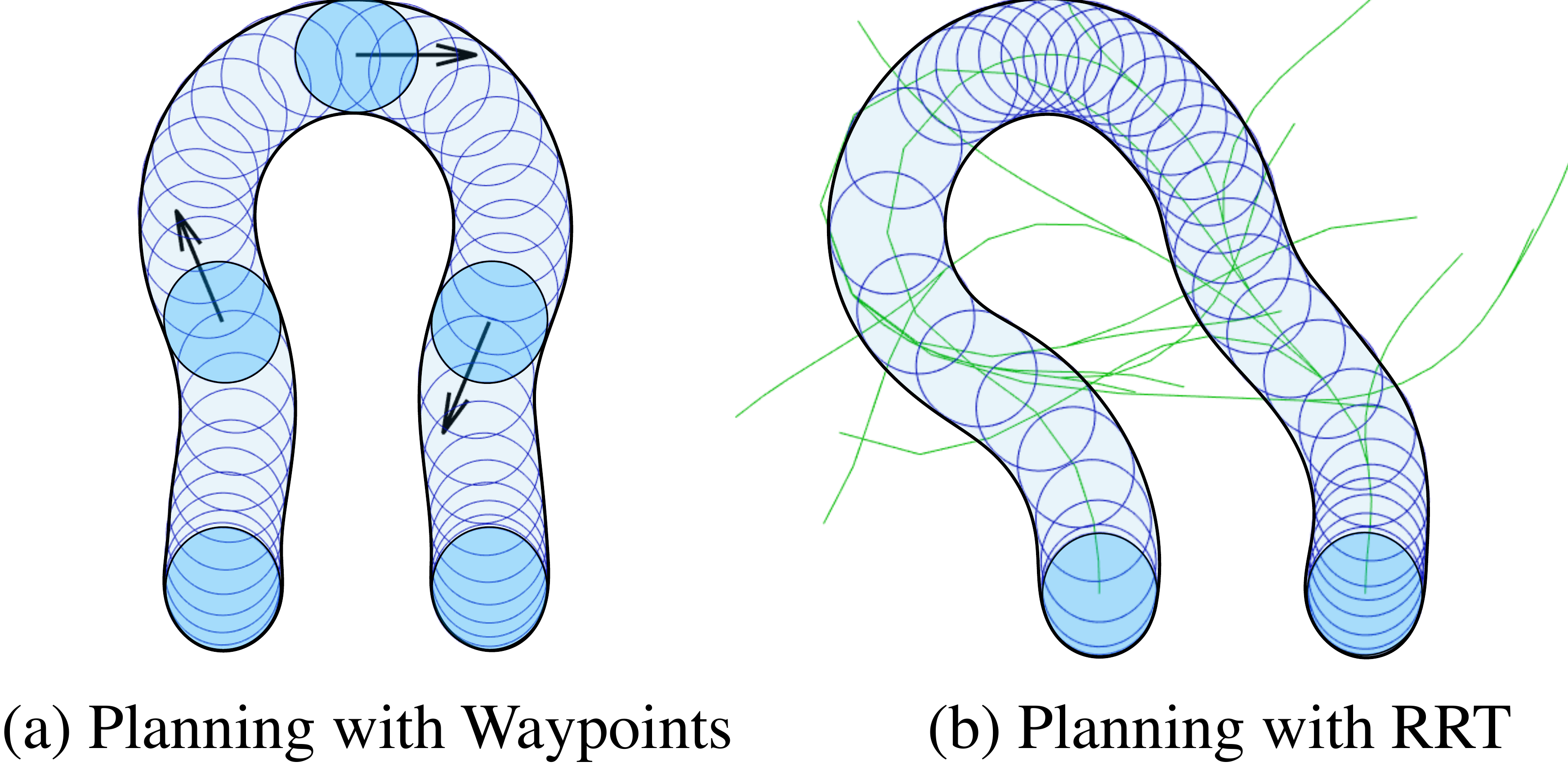}
\end{center}
\caption{Schematic View of Generated Funnels for Turning}\label{fig:turn} 
\end{figure}

\begin{figure}[t]
\vspace{0.2cm}
\begin{center}
	\includegraphics[width=0.4\textwidth]{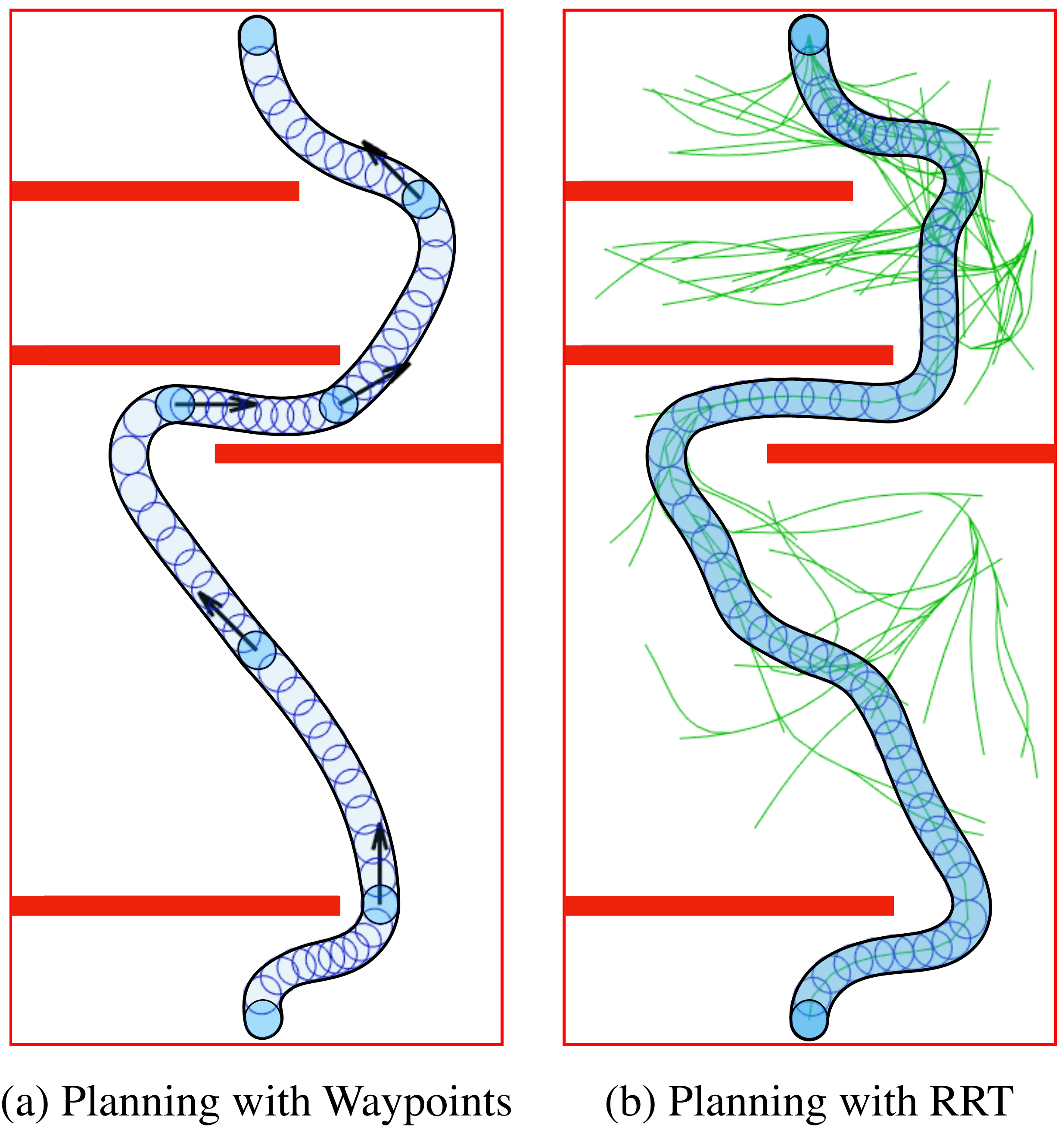}
\end{center}
\caption{Schematic View of Generated Funnels for Reach-Avoid}\label{fig:reach} 
\end{figure}

Suppose that the vehicle stopped at $[-0.5 \ 0 \ 0 \ 0]$ wants to reach $[0.5 \ 0 \ -\pi \ 0]$ (stopped position). While this is a hard task because of non-holonomic constraints, using a set of waypoints we can find a robust plan as shown in Fig.~\ref{fig:turn}(a). 
This can be extended to cases where there are obstacles in the system. Fig. ~\ref{fig:reach}(a), shows obstacles in a scene where the goal is to reach from $[0 \ -6 \ 0 \ 1]$ to $[0 \ 6 \ 0 \ 1]$ without stopping the car, while avoiding obstacles. Using waypoints and a trajectory optimization that accounts for collisions, we find a provably robust solution.

\paragraph{Motion Planning with Random Sampling}
Another approach is using random sampling instead of relying on a set of waypoints. Similar to LQR-trees~\cite{tedrake2010lqr}, we can use sampling methods to generate random funnel trees and explore reachable sets. As we are not performing any verification at runtime, the computation time is relatively fast. For example, for the case where the vehicle is avoiding obstacles, we ran RRT (equipped with our funnel generator) for $10$ rounds and in all cases it took less than $10$ seconds in MATLAB to find a provably robust plan. Random funnel trees generated using our funnel generator is shown in Fig.~\ref{fig:turn}(b) and Fig.~\ref{fig:reach}(b). Notice the tree is projected into a $2D$ plane.
Here, we emphasize that while sampling methods are known to be probabilistically complete under mild assumptions~\cite{tedrake2010lqr}, their performance depends significantly on the structure of the problem. \looseness = -1 

\section{Related Work}
Many previous works use stability/safety analysis to consider robustness of generated trajectories~\cite{ames2014,Manchester2018}. However, here we only discuss previous work which provide formally correct solutions.
Some recent studies address safe motion planning for specific types of systems. For example, assuming that the system can be controlled locally around any specific state, Singh et al.~\cite{ContractionMetrics} use contraction theory~\cite{Manchester2017} and Tube-MPC~\cite{LANGSON2004} to define a safety tube around any arbitrary trajectory. The main assumption in these work is that local invariant regions are small enough to address safety. \looseness = -1
  
In contrast to the aforementioned method, we use region of finite-time invariant where a safety tube is provided around a nominal trajectory for a finite time. These regions are calculated using different optimization techniques~\cite{TOBENKIN2011,Ahmadi2014,Althoff2014} and can be used \emph{after} a trajectory is generated.
Inspired by work on trajectory libraries~\cite{Stolle2006,sermanet2008learning}, Majumdar et al.~\cite{majumdar2017} propose to use a library of precomputed tubes. Then these tubes can be concatenated at runtime by considering compatibility constraints between different tubes to design a maneuver-automaton~\cite{Frazzoli2005}.

In this work, we use \names\ that can certify correctness for a continuum of trajectories. This is similar to the idea proposed by Majumdar et al.~\cite{majumdar2012} to use a parametrized family of tubes and at runtime one can choose proper parameters. More specifically, this method restricts the domain of funnel generator $\G$ to be a parametric polynomial reference $\vx_r^{\vp}(\cdot)$. For a given reference $\vx_r^{\vp}(\cdot)$ it finds a feedback law $\ctrl^{\vp}$ and a region of finite-time invariant using a function $V^{\vp}$ which are also parametrized by $\vp$. These computations are done off-line and at runtime one can select a proper parameter $\vp$ which yields a fixed reference trajectory and the corresponding feedback and region of finite-time invariant~\cite{majumdar2012}. The main issue is the fact that Majumdar et al.~\cite{majumdar2012} assume a set of feasible trajectories can be explicitly represented by parameterized polynomials, and these parameterized polynomials are available as the input of the procedure.
However, representing feasible trajectories using parameterized polynomials is not a trivial task which significantly limits the applicability of that method. 

Recently, Kousik et al.~\cite{kousik2017safe} provide an alternative approach using occupation measures. In this method only the feedback law and funnels are parametrized and the reference is not used inside the formulation.
In addition, they demonstrated that the whole motion planning can be performed by removing unsafe parameters with solving possibly expensive SDPs at runtime, while our method only provides a tool to be used inside different motion planning methods.
Moreover, these methods represent the whole funnel with parametrized polynomials up to a certain degree. In contrast, instead of parameterizing the reference, we only force constraints on the reference which allows to use structure of trajectories and produce more complicated (even non-polynomial) funnels.

Finally, Herbert et al.~\cite{FaSTrack} provide a solution for generating funnels for all possible trajectories. The method was also extended to use SOS programming~\cite{fastrak-sos}. As the planner is treated as an adversary, the method can be conservative in certain situations. To mitigate this, Fridovich-Keil et al.~\cite{fridovich2018planning} propose to use a library of motion planners with different models and input saturation level. Our work extends this work wherein not only input saturation level, but also trajectory domain is restricted. In other words, we propose to design a library of motion planner, each of which designed for a specific task.

\section{Conclusion}
In this work we extend FaSTrak~\cite{FaSTrack} to reduce tracking errors by restricting the reference generated by the motion planner. We provide an algebraic method based on SOS programming to find such funnel generators, and demonstrate effectiveness of our method on two case studies. 
In future, we wish to apply our method to larger systems using more scalable techniques~\cite{Ahmadi2014}. Also, to minimize experts effort, we wish to investigate techniques to systematically partition all trajectories into different classes and produce different funnel generators for each class.   
\bibliography{refs}
\bibliographystyle{plain}
\end{document}